\documentclass{article}

\pdfoutput=1
\usepackage{PRIMEarxiv}

\usepackage[utf8]{inputenc}
\usepackage[T1]{fontenc}   
\usepackage{hyperref}      
\usepackage{url}          
\usepackage{booktabs}       
\usepackage{amsfonts}      
\usepackage{nicefrac}      
\usepackage{microtype}     
\usepackage{diagbox}

\usepackage{array}
\usepackage{lipsum}
\usepackage{amsmath}
\usepackage{algorithm}  
\usepackage{algorithmicx}  
\usepackage{algpseudocode} 
\usepackage{amsmath} 
\usepackage{amssymb}
\usepackage{fancyhdr}      
\usepackage{graphicx}       
\usepackage{color}
\usepackage{natbib}
\graphicspath{{media/}}     

\newcommand{\bs}{{\boldsymbol s}}

\title{Pedestrian volume prediction using a Diffusion Convolutional Gated Recurrent Unit Model}

\author{
   Yiwei Dong \\
   School of Statistics \\
   Renmin University of China, China\\
   \texttt{ydong@ruc.edu.cn} \\
   \And
   Tingjin Chu \\
   School of Mathematics and Statistics \\
   The University of Melbourne, Australia\\
   \texttt{tingjin.chu@unimelb.edu.au} \\
  \AND
   Lele Zhang \\
  School of Mathematics and Statistics \\
  The University of Melbourne, Australia\\
  ARC Training Centre in Optimisation Technologies,\\ Integrated Methodologies, and Applications\\
  \texttt{lele.zhang@unimelb.edu.au} \\
  \And
     Hadi Ghaderi \\
  School of Business, Law and Entrepreneurship \\
  Swinburne University of Technology, Australia\\
  \texttt{hghaderi@swin.edu.au} \\
  \And
  Hanfang Yang \thanks{Corresponding Author} \\
  School of Statistics \\
  Renmin University of China, China\\
   \texttt{hyang@ruc.edu.cn} \\
}

\begin{document}
\maketitle

\begin{abstract}
Effective models for analysing and predicting pedestrian flow are important to ensure the safety of both pedestrians and other road users. These tools also play a key role in optimising infrastructure design and geometry and supporting the economic utility of interconnected communities. The implementation of city-wide automatic pedestrian counting systems provides researchers with invaluable data, enabling the development and training of deep learning applications that offer better insights into traffic and crowd flows. Benefiting from real-world data provided by the City of Melbourne pedestrian counting system, this study presents a pedestrian flow prediction model, as an extension of Diffusion Convolutional Grated Recurrent Unit (DCGRU) with dynamic time warping, named DCGRU-DTW. This model captures the spatial dependencies of pedestrian flow through the diffusion process and the temporal dependency captured by Gated Recurrent Unit (GRU). Through extensive numerical experiments, we demonstrate that the proposed model outperforms the classic vector autoregressive model and the original DCGRU across multiple model accuracy metrics. 
\end{abstract}

\noindent%

\section{Introduction}
\label{sec:intro}

Crowd and pedestrian management plays a crucial role in enhancing the urban environment, livability, and mobility.
Poorly managed pedestrian flow could result in sub-optimal use of public spaces and transport systems and also lead to adverse events such as fatalities and injuries \citep{Feng&Miller2014}. The number of crowd and pedestrian related disasters has been on the rise around the world for a variety of reasons, including a lack of timely information about special conditions and anomalies in crowd behaviour \citep{Feliciani&Nishinari2018}. Tools for analysis of pedestrian flows are essential for the planning and geometric aspects of cities; design of infrastructure and addressing crowd safety concerns \citep{Haghani2020}. Without such insight, identifying the appropriate crowd and pedestrian response strategy remains ineffective \citep{Martella2017}. Therefore, predicting accurate spatio-temporal information about pedestrian activities such as volume, velocity and direction is fundamental to public safety and efficient urban planning \citep{Zhang2022}.

The City of Melbourne has deployed a pedestrian counting system that benefits from thermal and laser based sensors. This non-vision based technology, allows for recording multi-directional pedestrian movements by forming a counting zone, without invading privacy \citep{Wang2017}.  In this study, we develop a deep learning based tool and apply it to this data for the assessment and prediction of pedestrian flows and patterns, which in turn can be used to identify anomalies and extreme conditions presented in pedestrian flow behaviour at a city-wide scale \citep{Zaki2018}.

There is an increasing interest in using deep learning for spatial and spatio-temporal data from statisticians. For spatial data, a deep compositional model is used to model the nonstationary and anisotropic covariance functions 
\citep{zammit2022deep, vu2022modeling}. Similarly, a deep neural network can be used to model 
the spatial mean trend \citep{zhan2024neural}. For spatio-temporal data, the echo state network, a type of deep neural network, is used in many applications, such as sea-surface temperature \citep{mcdermott2017ensemble}, soil moisture \citep{mcdermott2019deep}, wind power \citep{huang2022forecasting}, and air pollution \citep{bonas2023calibration}. A more detailed review of the statistical deep learning methods for spatial and spatio-temporal data can be found in \citet{wikle2023statistical}.

Urban traffic data pose a unique challenge; that is, the Euclidean distance, while still useful, may not be as useful as other applications such as air pollution or temperature forecasting. 
For traffic data,  graph neural networks (GNNs) have received additional attention because of their ability to handle complexities among nodes 
\citep{jiang2022graph, ye2020build, Wang2020, li2017diffusion}. More specifically, GNNs are ideal tools to model the spatial dependency between different sites, roads, or regions in traffic forecasting problems \citep{jiang2022graph}. Typically in GNNs, the traffic graph is constructed as a triple consisting of the node-set, the edge set, and the adjacency matrix. As the adjacency matrix in GNNs plays an important role in capturing spatial dependencies, various ways are proposed to represent such a matrix in traffic forecasting problems \citep{jiang2022graph}. For a road based matrix, the road connection matrices, transportation connectivity matrices, and direction matrices are merged for modelling complex road networks such as arterial roads, highways, and subway systems effectively \citep{ye2020build}. The distance-based matrix concentrates on the spatial closeness between nodes, and is used in \citet{lee2022ddp} to model the distance, direction and positional relationship. Another type of adjacency matrix is the similarity based matrix, which considers traffic patterns or functional similarity matrices to reveal the intrinsic mutual dependencies, correlations and distributions of different locations \citep{jiang2021dl}. Dynamic matrices are also considered in traffic forecasting tasks as an alternative to pre-defined adjacency matrices \citep{bai2020adaptive, XU2023580, Gu2023, Han2021}.

Compared to motor vehicle related modelling, there are fewer works in the literature dedicated to pedestrian or crowd flow forecasting. We identified three groups of related studies: (i) passenger flow prediction at public transportation stations such as rail transit stations or bus stations \citep{9257011, li2017, Liu2019, peng2020spatial}, (ii) crowd flow prediction using GPS data collected from cell phones, taxis, shared bikes, etc. \citep{Zhang2020, Zhang2022, Yuan2020, Duives2019}, and (iii) pedestrian volume prediction using the data collected by various types of counting systems such as camera or thermal sensors across a large region \citep{liu2021pedestrian}. The focus of this work is (iii). 
Although the aforementioned models, as well as other traffic forecasting approaches can be used for pedestrian or crowd flow forecasting, we acknowledge that pedestrian flow forecasting problems represent unique complexities \citep{liu2021pedestrian}. Unlike motor vehicles, whose trajectories and velocities are restricted by road network conditions and traffic regulations, pedestrian flows and patterns often present significant uncertainties \citep{8578651}. As such, in many cases, it is naturally more difficult to simply employ traffic forecasting models developed for motor vehicles for pedestrian flow prediction and various special features of pedestrian flow need to be considered when designing the model \citep{liu2021pedestrian, doi:10.1080/15568318.2020.1858377}.

In this paper, we present a Diffusion Convolutional Gated Recurrent Unit (DCGRU) enhanced with Dynamic Time Warping (DTW) for pedestrian volume forecasting. To address the challenge of measuring proximity between locations—given the freedom pedestrians have in choosing their paths—we leverage DTW, a robust time series similarity measure, as additional information in constructing the adjacency matrix of the graph neural network within the DCGRU. By doing so, our approach effectively captures the nuanced spatio-temporal relationships in pedestrian flow data. The proposed DCGRU-DTW model outperforms four other methods, including the original DCGRU and the classic VAR model, demonstrating its effectiveness in accurately forecasting pedestrian volumes across various scenarios.

The remainder of the paper is organized as follows. 
In Section~\ref{sec:omodel}, we introduce the components of the original DCGRU architecture proposed by \citet{li2017diffusion}. Our proposed DCGRU with dynamic time warping model is presented in Section~\ref{sec:model}. In Section~\ref{sec:data}, the pedestrian volume of the City of Melbourne is analysed and results are presented.

\section{An introduction to DCGRU Model} \label{sec:omodel}

In this section, we first formulate the pedestrian volume forecasting problem, and then the components of original DCGRU \citep{li2017diffusion} is introduced.

\subsection{Pedestrian Volume Forecasting Problem Formulation}

We consider a system of sensors deployed at intersections in an urban area to measure pedestrian volumes. 
The pedestrian volume forecasting problem in this study is aimed to predict the number of pedestrians passing through each sensor in the future. Let $X_{i, t}$ be the observed number of pedestrians detected at the $i$-th sensor during the timestamp $t$ for $t=1,2,\dots, T$. 
The size of a timestamp in this study is one hour. 
The collection of the data from all the sensors forms a multivariate time series $\{\boldsymbol{X}_1, \ldots,  \boldsymbol{X}_{T}\}$, where $\boldsymbol{X}_t = (X_{1, t}, \ldots, X_{N, t})^T \in \mathbb{R}^{N}$ and $N$ is the number of sensors. 
The goal of the pedestrian volume forecasting problem is to seek a function $h(\cdot)$ that maps the data of the past $T$ timestamps to predict the future values of $T'$ timestamps, given some other information $\mathcal{I}$ such as the geographical information in the traffic study:
$$
\left[ \boldsymbol{X}_1, \cdots, \boldsymbol{X}_{T} ; \mathcal{I} \right] \stackrel{h(\cdot)}{\longrightarrow}\left[\boldsymbol{X}_{T+1}, \cdots, \boldsymbol{X}_{T+T'}\right].
$$

\subsection{Gated Recurrent Unit}

Gated Recurrent Unit (GRU) is one type of recurrent neural networks that shows excellent ability in modeling the temporal dynamics of sequential data \citep{chung2014empirical}. The input and output in each timestamp of the GRU is the same as that of the vanilla RNN \citep{medsker2001recurrent}: at each timestamp, there is an input denoted as $\boldsymbol{X}_t$, and the hidden state denoted as $\boldsymbol{H}_{t-1}$, which is passed from the previous state of $t-1$. The previous hidden state $\boldsymbol{H}_{t-1}$ contains relevant information from the previous time. Using the input $\boldsymbol{X}_t$ and the previous hidden state $\boldsymbol{H}_{t-1}$, the current hidden state $\boldsymbol{H}_t$ is computed by the GRU cell and then passed forward. Additionally, an output denoted as $\boldsymbol{Y}_t$ for the current timestamp is optionally produced by the GRU.

The inner structure of one GRU cell can be summarized as follows. {There are two gates in each GRU cell: reset and update.} The $t$-th states of the two gates are computed according to the following formulas:
\begin{equation}
   \begin{aligned}
\boldsymbol{r}_t&=\sigma\left(\boldsymbol{W_r} [\boldsymbol{X}_t,  \boldsymbol{H}_{t-1}]+\boldsymbol{b_r}\right)\\
\boldsymbol{z}_t&=\sigma\left(\boldsymbol{W_z} [\boldsymbol{X}_t,  \boldsymbol{H}_{t-1}]+\boldsymbol{b_z}\right)
\end{aligned} 
\end{equation}
where $\boldsymbol{r}_t$ denotes the state of the reset gate and $\boldsymbol{z}_t$ denotes the state of the update gate. Parameters $\boldsymbol{W}$ and $\boldsymbol{b}$ with different subscripts are learnable weights and biases, respectively of the two gates. Operator $[\cdot,\cdot]$ denotes the concatenation of matrices. For example, $[\boldsymbol{X}_t,  \boldsymbol{H}_{t-1}]$ means a new matrix containing both $\boldsymbol{X}_t$ and $\boldsymbol{H}_{t-1}$. The gating mechanism is used to control the information flow. 
The reset gate is responsible for the short-term memory of the network, determining how much past information to forget. It is used to calculate the current candidate activation $\boldsymbol{c}_t$:
\begin{equation}
    \boldsymbol{c}_t=\tanh \left(\boldsymbol{W_c}[ \boldsymbol{X}_t, \left(\boldsymbol{r}_t \odot \boldsymbol{H}_{t-1}\right)]+\boldsymbol{b_c}\right),
\label{ct01}
\end{equation}
where $\odot$ is the element-wise multiplication and $\boldsymbol{W_c}$ and $\boldsymbol{b_c}$ are again weights and biases respectively. As we can infer from Equation~\eqref{ct01}, the closer $\boldsymbol{r}_t$ is to $0$,  the less information of the previous state would be included in the current candidate activation. When the reset gate is off, i.e., $\boldsymbol{r}_t$ approximates $0$, then nearly all the previous information is forgotten by $\boldsymbol{c}_t$. The update gate $\boldsymbol{z}_t$ is responsible for the long-term memory of the network. It controls how much information to carry forward from the previous hidden state $\boldsymbol{H}_{t-1}$ and also controls how much new information from the current candidate activation needs to be added to the current hidden state $\boldsymbol{H}_t$. The current hidden state is then computed by:
\begin{equation}
\boldsymbol{H}_t=\boldsymbol{z}_t \odot \boldsymbol{H}_{t-1}+ \left(1-\boldsymbol{z}_t\right) \odot \boldsymbol{c}_{t}.
\end{equation}

\begin{figure}[tbp]
    \centering
    \includegraphics[scale=0.56]{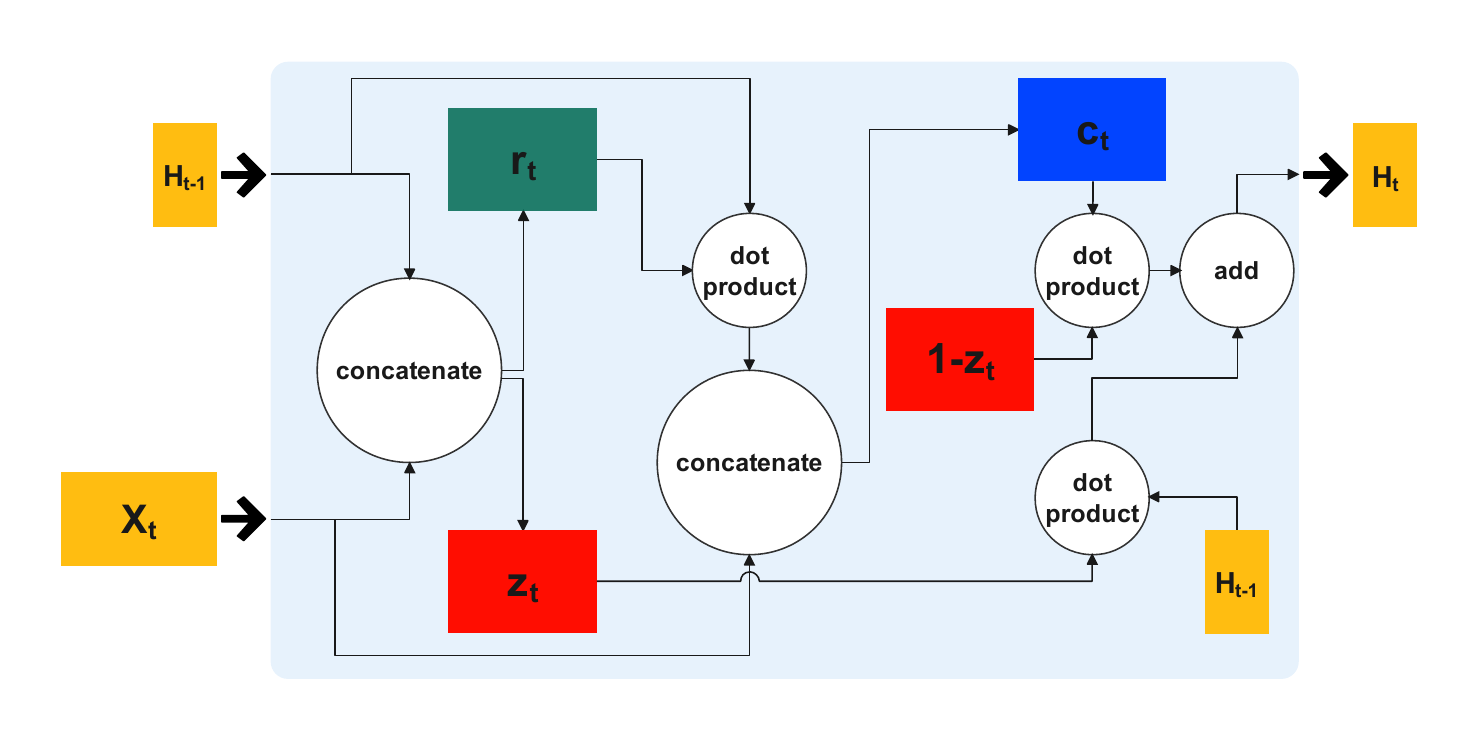}
    \caption{The inner structure of one GRU cell}
    \label{fig:grucell}
\end{figure}

Accordingly,  the GRU cell finishes the entire calculation process of time $t$ via the gating mechanism. In a GRU network composed of multiple stacked cells, the individual cells exhibit specialized functions: some cells with more active reset gates ($\boldsymbol{r}_t$ is close to $1$) are better suited to capturing short-term dependencies, while others, where the update gate is more active ($\boldsymbol{z}_t$ is close to $1$), focus on capturing long-term dependencies. Consequently, the GRU network, through the combined efforts of its cells, is capable of adaptively modeling both short-term and long-term temporal dependencies. The illustration of the inner structure of one GRU cell is depicted in Figure~\ref{fig:grucell}. 

\subsection{Diffusion Convolutional Layer}

Modelling the spatial dependency of pedestrian volumes at different locations is challenging and has not been well investigated \citep{PFIESTER2021}. 
The diffusion convolution, as well as the corresponding diffusion convolutional layer, which utilizes bidirectional random walks on the graph, is an effective model to capture the spatial dependency on road networks \citep{li2017diffusion}. Concretely, the sensor network is represented as a graph where the nodes represent the sensors and the edges represent their proximity. Formally, the sensor network graph $\mathcal{G}=(\mathcal{V}, \mathcal{E}, \boldsymbol{\mathcal{W}})$, where $\mathcal{V}$ is a set of nodes whose number is equal to the number of sensors $|\mathcal{V}|=N$, $\mathcal{E}$ is a set of edges and $\boldsymbol{\mathcal{W}} \in \mathbb{R}^{N \times N}$ is a weighted adjacency matrix. Here, $\boldsymbol{\mathcal{W}}$ is calculated based on predefined road network distances and is typically not symmetric because of one-way streets and traffic control. The traffic flow data observed on this graph at time $t$ is called the graph signal, which is denoted as $\boldsymbol{X}_t \in \mathbb{R}^{N \times P}$, where $P$ is the number of features of each node. The diffusion convolution operation on this graph signal is then defined using a filter  $f_{\boldsymbol{\theta}}$ that is parameterized by the parameter $\theta$:

\begin{equation}
   \boldsymbol{X}_{t}[:, p] \star_{\mathcal{G}} f_{\boldsymbol{\theta}}=\sum_{k=0}^{K-1}\left(\theta_{k, 1}\left(\boldsymbol{D}_O^{-1} \boldsymbol{\mathcal{W}}\right)^k+\theta_{k, 2}\left(\boldsymbol{D}_I^{-1} \boldsymbol{\mathcal{W}}^{\top}\right)^k\right) \boldsymbol{X}_{t}[:, p]\quad \text { for }  p \in\{1, \cdots, P\}, 
   \label{dfo}
\end{equation}
where $\boldsymbol{D}_O=\operatorname{diag}(\boldsymbol{\mathcal{W}} \mathbf{1})$ is the out-degree diagonal matrix and $\mathbf{1} \in \mathbb{R}^N$ denotes a vector of $1$'s. The matrix  $\boldsymbol{D}_O^{-1} \boldsymbol{\mathcal{W}}$ in Equation~\eqref{dfo} is a state transition matrix of the diffusion process, where $\theta_{k, 1}$ is the corresponding parameter of $f_{\boldsymbol{\theta}}$ to be learned, and $\boldsymbol{D}_O^{-1}$ can be regarded as a normalizer for weighted adjacency matrix $\boldsymbol{\mathcal{W}}$. Similarly, $\boldsymbol{D}_I=\operatorname{diag}(\boldsymbol{\mathcal{W}}^T \mathbf{1})$ is the in-degree diagonal matrix and $\theta_{k, 2}$ is the parameter related to reversed diffusion process. When $\boldsymbol{\mathcal{W}}$ is symmetric, $\boldsymbol{D}_O$ and $\boldsymbol{D}_I$ become the same and can be merged to one. Constant $K$ is the maximum diffusion step. The essence of $\left(\boldsymbol{D}_O^{-1} \boldsymbol{\mathcal{W}}\right)^k \boldsymbol{X}_{t}[:, p]$ is, after the $k$-step diffusion, the information contained in related nodes is aggregated to each node according to their relevance, which is quantified by the probability of transition between nodes.

The diffusion convolutional layer employs the diffusion convolution operation defined in Equation~\eqref{dfo} to conduct further transformation to the original $P$-dimensional features. It maps the input $\boldsymbol{X}_t \in \mathbb{R}^{N \times P}$ to the output $\boldsymbol{O}_t \in \mathbb{R}^{N \times Q}$ by using filters $\left\{f_{\boldsymbol{\Theta}_{q, p,:},:}\right\}$ whose parameters are $\boldsymbol{\Theta}_{q, p,:,:}$, and an activation function $\boldsymbol{a}$ such as ReLU \citep{agarap2018deep} and Sigmoid \citep{han1995influence}:
\begin{equation}
    \boldsymbol{O}_t[:, q]=\boldsymbol{a}\left(\sum_{p=1}^P \boldsymbol{X}_t[:, p] \star_{\mathcal{G}} f_{\boldsymbol{\Theta}_{q, p, i,:}}\right) \quad \text { for } q \in\{1, \cdots, Q\}
\end{equation}
The diffusion convolutional layer has demonstrated effectiveness in spatial data feature extraction. In the following section, we use $\boldsymbol{\Theta} \in \mathbb{R}^{Q \times P \times K \times 2}$ to denote the collection of all parameters in the diffusion convolutional layer.

In this study, we adopt the diffusion convolutional layer to model the spatial dependency of pedestrian counts at various locations, i.e. $P=1$. For road traffic networks, distances and the adjacency matrix $\boldsymbol{\mathcal{W}}$ between sensors are relatively well-defined as vehicles travel on roads and follow traffic rules. In contrast, pedestrians could travel freely in arbitrary direction and is less controllable, making it difficult to define distances and the adjacency matrix. The next section explains how $\boldsymbol{\mathcal{W}}$ is computed for pedestrian sensor networks.

\section{DCGRU Model with Dynamic Time Warping} \label{sec:model}

In this section, we present the modified DCGRU model for pedestrian volume prediction. 
Compared with the original DCGRU \citep{li2017diffusion}, the modified DCGRU improves the spatio-temporal forecasting ability by incorporating the connectivity and proximity between pedestrian counting sensors via dynamic time warping.

\subsection{Dynamic Time Warping}

The adjacency matrix $\boldsymbol{\mathcal{W}}$ has a great impact on the performance of graph neural networks. Geographic information, such as latitude and longitude, is commonly used to construct the adjacency matrix. However, in the field of transportation (both vehicular and pedestrian), the geographic information alone may not be sufficient to capture the relationships between different sensors \citep{liu2021pedestrian, Hu2021}. Despite two sensors being geographically closer, the time series of pedestrian volumes at these two location may present completely different pattern. An example is given in the supplementary material. 
Therefore, in order to better capture the relationship among pedestrian counting sensors, this study proposes to incorporate the distance of two time series, besides the distance based on geographical information to calculate a weighted adjacency matrix 
\begin{equation}
    \boldsymbol{\mathcal{W}} = \boldsymbol{\mathcal{W}}_{geo} + \beta \boldsymbol{\mathcal{W}}_{ts},
\end{equation}
where $\boldsymbol{\mathcal{W}}_{geo}$ is calculated based on geographical information of sensors, $\boldsymbol{\mathcal{W}}_{ts}$ is calculated based on time series using DTW, and $\beta$ is a hyper-parameter that controls how much $\boldsymbol{\mathcal{W}}_{ts}$ contributes to the final weighted adjacency matrix.

DTW is a well-established time series distance, and it can effectively deal with scaling and translation on the time axis. Intuitively speaking, for a point in a time series, DTW can match the points around the corresponding position of the point in another time series. In this way, the time axis can be viewed as stretchable, rather than one-to-one like the traditional Euclidean distance. A formal definition of the DTW distance is given as follows.

Let $X = \{x_i\}_{i=1}^N$ be a time series of length $N$ and $Y = \{y_j\}_{j=1}^ M$ be another time series of length of $M$,  where $N$ is not necessarily equal to $M$. We denote that the local distance between the $i$-th element $x_i$ in $X$ and the $j$-th element $y_j$ in $Y$ is $c(x_i, y_j)$. Here,  $c$ is a metric defined in the space $\mathcal{F}$ where $x_i$ and $y_j$ locate (assumed to be the same),  $ c:\mathcal{F} \times \mathcal{F} \rightarrow \mathbb{R} $. Define the cost matrix $C\in \mathbb{R}^{N \times M}$ as $C(i,  j) = c(x_i,  y_j)$. Additionally, the warping path can be expressed as a sequence of pairs $(p_1, \ldots, p_L)$, where each $p_l = (i_l, j_l)$ belongs to the Cartesian product of the intervals $[1:N]$ and $[1:M]$, and $L$ represents the number of steps, which is not predetermined. The sequence $p$ must fulfil three conditions: (1) the boundary condition requires $p_1 = (1,1)$ and $p_L = (N, M)$, (2) the monotonicity condition demands that $ i_1\leqslant i_2 \leqslant \ldots\leqslant i_L$, while $ j_1 \leqslant j_2\leqslant\ldots
   \leqslant j_L$, and (3) the step size condition requires that  $p_{l+1}-p_l$, i.e. the difference between two consecutive pairs of elements in $p$ can only be one of the following three tuples: $(1,0)$, $(0,1)$, or $(1,1)$, for $l$ in the range $[1:L-1]$. Then, the total cost of a warping path $p$ between time series $X$ and $Y$ is given by the sum of the costs of each point in the path, calculated as $c_p(X, Y) = \sum_{l=1}^L c(x_{i_l}, y_{j_l})$. Finally, the DTW distance between $X$ and $Y$ is defined as the minimum of the total costs of all possible warping paths, which can be expressed as:
          \begin{equation}
           \text{DTW}(X, Y)=\operatorname{min}\{c_{p}(X, Y)\ |\ p \text{ is a warping path} \}\text{. }
       \end{equation}

\subsection{Sequence to Sequence Framework}
\label{subsec:seq2seq}

Sequence to sequence (Seq2Seq) framework is an end-to-end deep learning framework \citep{sutskever2014sequence}, and it can effectively tackle the multi-step to multi-step sequence learning problems. Seq2Seq consists of an encoder and a decoder. In most cases, both the encoder and decoder are RNNs. The encoder encodes the input sequence into a latent vector,  which is referred to as {context}, and then the context is decoded by the decoder to yield outputs. During the decoding process, the decoder recurrently leverages the true label (during training) or output of the last timestamp (during inference) as the input of the current timestamp, and performs the decoding operation to yield the new hidden state and output until the stop symbol is the output. Compared with simple RNNs, including those with a gated mechanism, the Seq2Seq framework generally performs better in long-range time series forecasting \citep{LINDEMANN2021650} and can deal with variable-length input and output.

\subsection{Diffusion Convolutional Gated Recurrent Unit Model}

\begin{figure}[t]
    \centering
    \includegraphics[width=\textwidth]{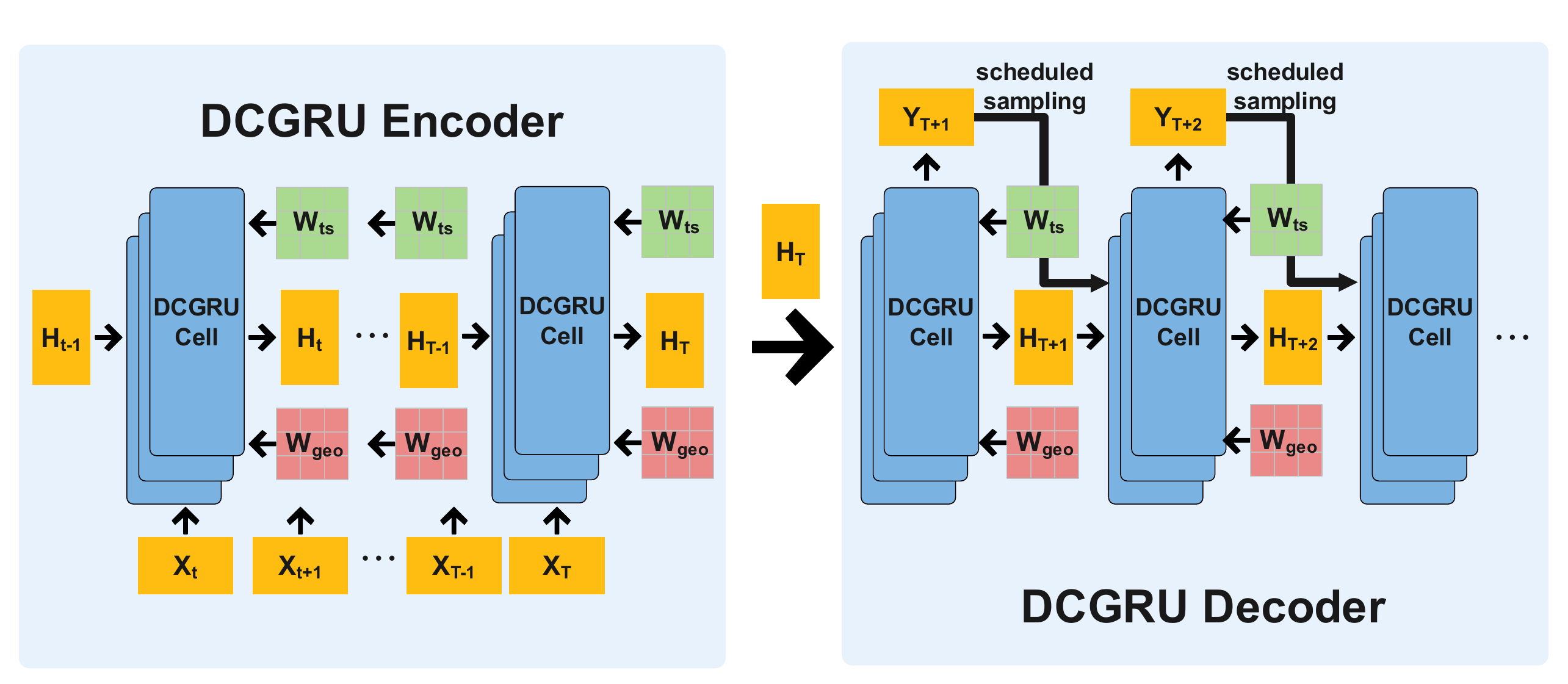}
    \caption{The overall framework of DCGRU}
    \label{fig:dcgru}
\end{figure}

Leveraging on the aforementioned model components and techniques, we now explain the overall architecture of the diffusion convolutional gated recurrent unit (DCGRU) model for the pedestrian volume forecasting problem, as illustrated in Figure~\ref{fig:dcgru}. 
The DCGRU also benefits from the Seq2Seq framework, and each layer in both the encoder and decoder is a DCGRU cell, whose inner structure is described as:
\begin{align}
    &\boldsymbol{r}_{t} =\sigma\left(\boldsymbol{\Theta}_r \star_{\mathcal{G}}\left[\boldsymbol{X}_{t},  \boldsymbol{H}_{t-1}\right]+\boldsymbol{b}_r\right) \label{eq9}\\
    &\boldsymbol{z}_{t}=\sigma\left(\boldsymbol{\Theta}_z \star_{\mathcal{G}}\left[\boldsymbol{X}_{t},  \boldsymbol{H}_{t-1}\right]+\boldsymbol{b}_u\right) \label{eq10}\\
&\boldsymbol{c}_{t}  =\tanh \left(\boldsymbol{\Theta}_c \star_{\mathcal{G}}\left[\boldsymbol{X}_{t}, \left(\boldsymbol{r}_{t} \odot \boldsymbol{H}_{t-1}\right)\right]+\boldsymbol{b}_c\right)\label{eq11}\\
&\boldsymbol{H}_{t}=\boldsymbol{z}_{t} \odot \boldsymbol{H}_{t-1}+\left(1-\boldsymbol{z}_{t}\right) \odot \boldsymbol{c}_{t}\label{eq12}
\end{align}
where $\boldsymbol{X}_t$ are the input and $\boldsymbol{H}_{t-1}$ is the hidden state of timestamp $t$. $\Theta_r \star_{\mathcal{G}}$, $\Theta_z \star_{\mathcal{G}}$ and $\Theta_c \star_{\mathcal{G}}$ are diffusion convolution operations for the corresponding filters. By Equation \eqref{eq9} - \eqref{eq12}, the DCGRU cell substitutes the linear transformations in the original GRU with diffusion convolutions, and thus achieves the spatiotemporal feature extraction. Moreover, several DCGRU cells can be stacked in the encoder and the decoder of Seq2Seq to obtain better feature extraction results.

Scheduled sampling \citep{10.5555/2969239.2969370} is used to mitigate the cumulative error problem, which is caused by the discrepancy between training and inference. Specifically, during the inference phase of standard Seq2Seq models, the output of each timestamp has an association with the outputs of previous timestamps. If a wrong output is generated, the input state of all the following timestamps will be affected, and the error will continue to accumulate. Scheduled sampling alleviates this problem by no longer completely using true labels as inputs at each timestamp during training. Instead, the true label is fed into the decoder with a probability $\epsilon_i$ at $i$-th iteration, and the output of the model itself is chosen to be fed into the decoder with probability $1 - \epsilon_i$.

\section{Melbourne Pedestrian Volume Forecasting}
\label{sec:data}

\subsection{Some Details on Data Preprocessing} \label{sec:datapre}

To determine variations in pedestrian activity throughout the day, the City of Melbourne has built a pedestrian counting system, which is composed of pedestrian counting sensors installed across the central city area of Melbourne.
The dataset records the hourly counts of pedestrians passing by various sensors in the City of Melbourne and is available online ({\url{http://www.pedestrian.melbourne.vic.gov.au}). 
Locations of sensors can be obtained via the Sensor Locations website ({\url{https://data.melbourne.vic.gov.au/explore/dataset/pedestrian-counting-system-sensor-locations/information}). 
See Figure~\ref{fig:sensor_map}. In order to avoid the impact of the COVID-19 pandemic and the consequential lockdowns, we selected the data from April 1 to December 31 of 2019. There are missing values in the dataset due to various reasons, such as sensor malfunction. We selected $30$ sensors with the least number of missing values for further analysis. For the missing values in the data of these $30$ sensors, we chose the average of the non-missing values of the same sensor from the same hour to fill in. 

\begin{figure}[htbp]
    \centering
    \includegraphics[width=0.9\textwidth]{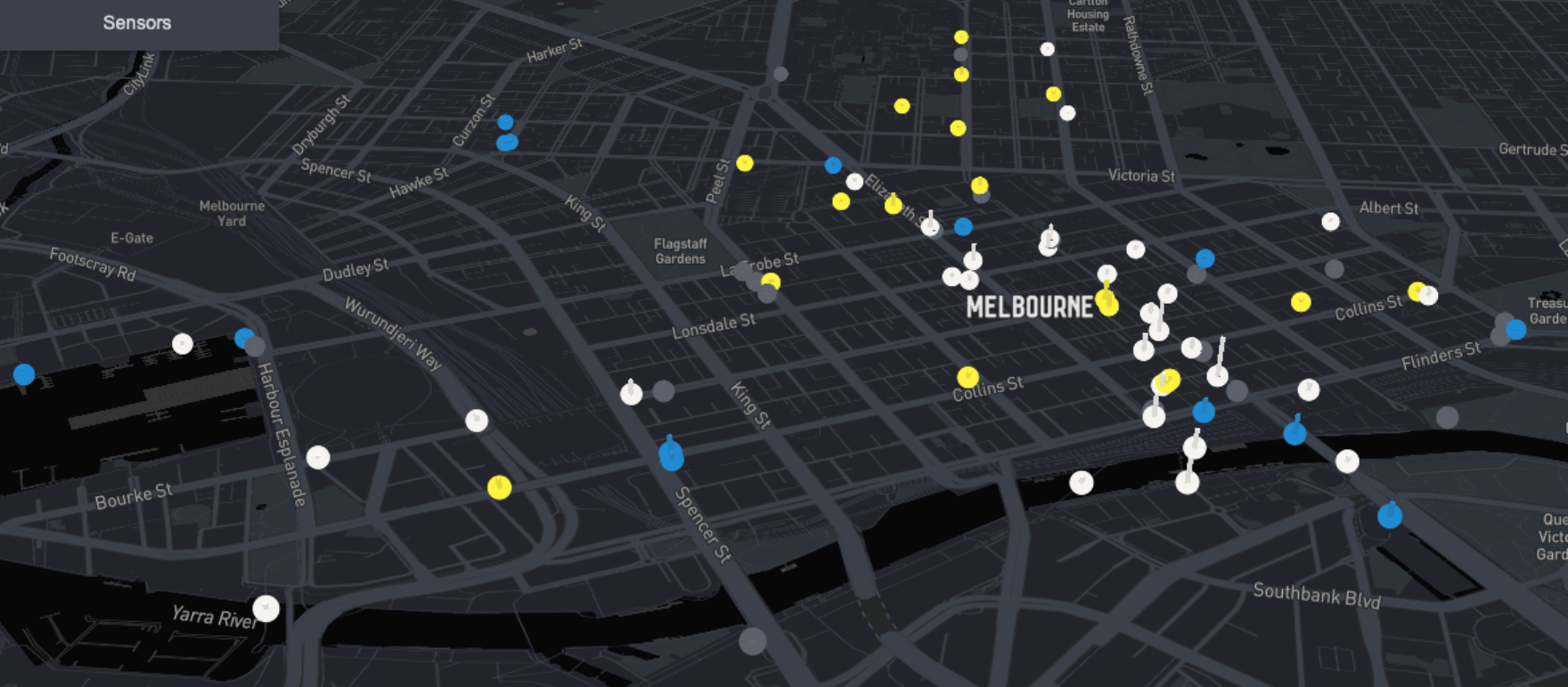}
    \caption{The sensor map of the pedestrian counting system in Melbourne}
    \label{fig:sensor_map}
\end{figure}

\begin{figure}[t]
    \centering
    \includegraphics[width=0.75\textwidth]{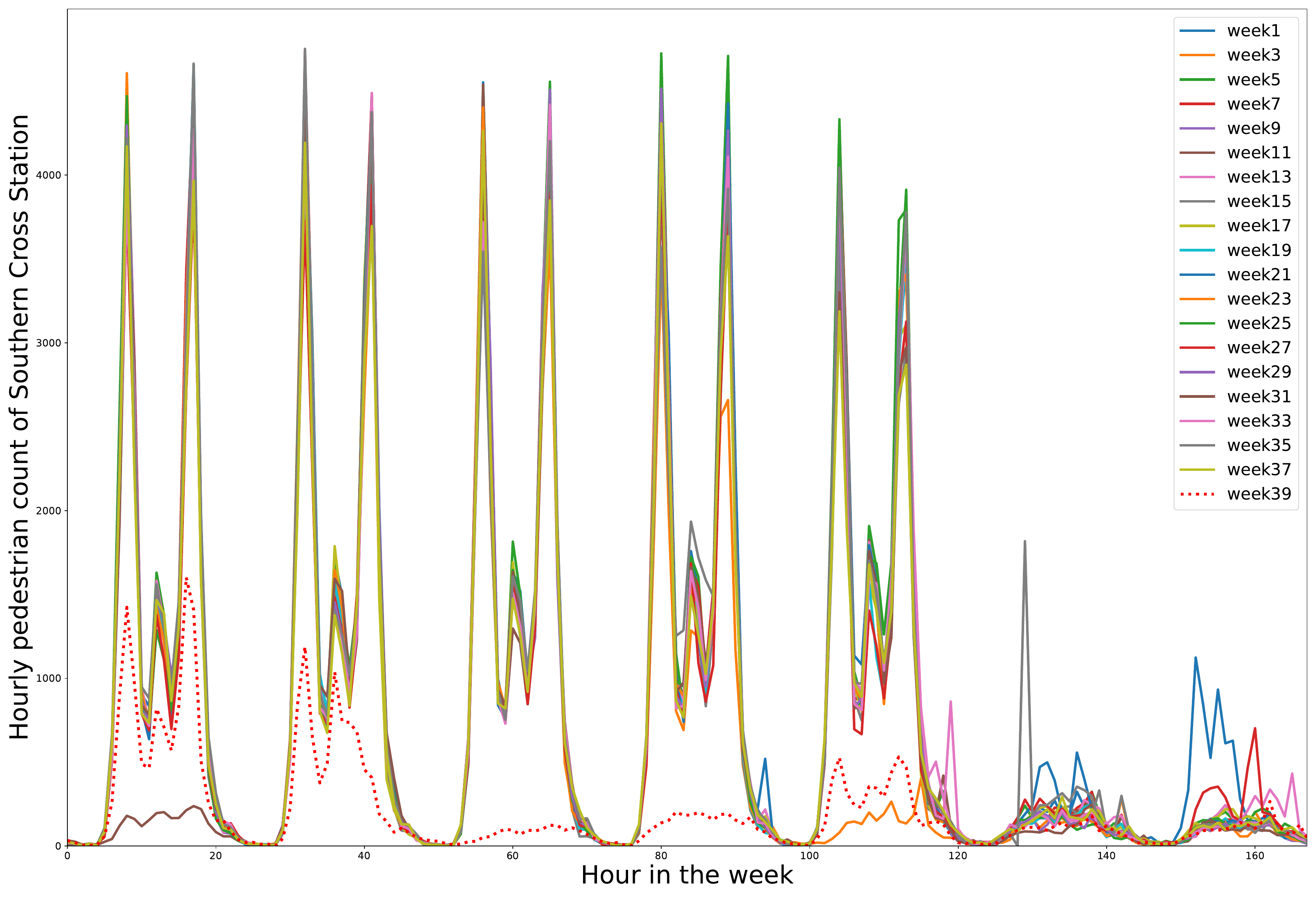}
    \caption{Hourly pedestrian counts at Southern Cross Station in different weeks. The data for week 39 (from December 23 to December 29) marked as a dotted line, shows a different pattern compared with the other weeks.}
    \label{fig:abn}
\end{figure}

Besides missing data, our exploratory analysis also identified abnormal data. We found that the data of certain weeks present very different characteristics from others; an example is shown in Figure~\ref{fig:abn}. This discrepancy may be due to factors such as the Christmas holidays or extreme weather. To unveil the general spatio-temporal patterns of pedestrian volumes, we applied an anomaly detection algorithm based on $k$-medoids clustering to eliminate the influence of data in abnormal weeks. For more details, see the supplementary material.

\subsection{Numerical Experiment Setup}

In this section, we describe the implementation details of the DCGRU model and experiment settings. We use the first 70\% of the data in chronological order as the training set, the next 10\% as the validation set, and the last 20\% of the data as the test set. Similar to most time series supervised learning tasks, the sliding window method is used to construct the input data of the DCGRU model: let $L_{input}$ denote the input length of each window and $L_{output}$ denote the output length. Then, for the current timestamp $t$, the input multivariate time series data of the DCGRU model is $
\left[ \boldsymbol{X}_{t-L_{input}+1}, \cdots, \boldsymbol{X}_{t}\right]$ and the expected output is $\left[\boldsymbol{X}_{t+1}, \cdots, \boldsymbol{X}_{t+L_{output}}\right]$. In the experiment, we keep $L_{output} = 5$ and the step of sliding window be $1$. The input length $L_{input}$ is set to $5$ and $168$ (one week), and the performance of the two input lengths is evaluated.

For the adjacency matrix, $\boldsymbol{\mathcal{W}}_{geo}$ is constructed by using sensors' longitude and latitude coordinates. More specifically, the geographical adjacency matrix is calculated via a thresholded Gaussian kernel: 
\begin{equation}
\boldsymbol{\mathcal{W}}_{geo}(i, i')=\left\{ 
\begin{array}{ll}
\exp \left(-\frac{{d}\left(\bs_i, \bs_{i'}\right)^2}{\sigma_g^2}\right), & \quad \mbox{if}\quad {d}\left(\bs_i, \bs_{i'}\right) \leqslant \kappa, \\ 
0, & \quad \mbox{otherwise},  
\end{array}%
\right. \newline
\end{equation}
where $\kappa$ is a threshold, $\sigma_g$ is the standard deviation of all the geographical distances, $\bs_i$ and $\bs_{i'}$ denote the coordinates of the sensors $i$ and $i'$, respectively. The thresholded Gaussian kernel can make the adjacency matrix sparse, which benefits the computation of the diffusion convolution operation. 
The thresholded Gaussian kernel is also used to compute $\boldsymbol{\mathcal{W}}_{ts}(i, i')$.  We select a representative period to calculate DTW distances between sensors to reduce the computation complexity. That is, 
\begin{equation}
\boldsymbol{\mathcal{W}}_{ts}(i, i')=\left\{ 
\begin{array}{ll}
\exp \left(-\frac{\text{DTW}(\boldsymbol{c}_{i}, \boldsymbol{c}_{i'})^2}{\sigma_t^2}\right), & \quad \mbox{if}\quad \text{DTW}(\boldsymbol{c}_{i}, \boldsymbol{c}_{i'}) \leqslant \kappa, \\ 
0, & \quad \mbox{otherwise},  
\end{array}%
\right. \newline
\end{equation}
where $\boldsymbol{c}_{i}$ is the $i$-th column of the cluster centre derived in Section \ref{sec:datapre}, which represents the pattern of typical weekly pedestrian counts of the $i$-th sensor, $\sigma_t$ is the standard deviation of all the time series distances.  In our experiment, the sample standard deviation is used for $\sigma_g$ and $\sigma_t$, and the threshold $\kappa$ is set to be $0.1$.

Three evaluation metrics are used to quantify the performance of different models: 1) Mean Absolute Error (MAE), 
$
\operatorname{MAE}(\boldsymbol{X}, \hat{\boldsymbol{X}})=\frac{1}{NL}\sum_{t=t_0}^{t_0+L}\sum_{i=1}^N{ \left|\hat{X}_{i,t} - X_{i,t}\right|}
$, 2) Mean Absolute Percentage Error (MAPE), 
$
\operatorname{MAPE}(\boldsymbol{X}, \hat{\boldsymbol{X}})=\frac{1}{NL}\sum_{t=t_0}^{t_0+L}\sum_{i=1}^N{ \left|\frac{\hat{X}_{i,t} - X_{i,t}}{X_{i,t}}\right|} 
$, and 3)
Root Mean Square Error (RMSE), 
$
\operatorname{RMSE}(\boldsymbol{X}, \hat{\boldsymbol{X}})=\sqrt{\frac{1}{NL} \sum_{t=t_0}^{t_0+L}\sum_{i=1}^N \left(\hat{X}_{i,t} - X_{i,t}\right) ^2}
$, where $N$ is the number of sensors, $L$ is the length of the data to be tested, $t_0$ is the start timestamp of the test data, and $X_{i,t}$ and $\hat{X}_{i,t}$ denote the true and the predicted pedestrian volumes observed by sensor $i$ at time $t$, respectively.

For hyper-parameters tuning, we used the grid search method and repeated the experiment 3 times for each set of parameters. The range of weight $\beta$ is chosen from the set $\{0.01, 0.02, 0.05, 0.1, 0.2, 0.5, 1, 2, 5\}$; The maximum
diffusion step $K$ is $\{1, 2, 3\}$. The range of the learning rate in the Adam optimizer is $\{0.001, 0.005, 0.01\}$; The batch size is either 32 or 64; The number of layers of the DCGRU cell is 1 or 2. For each set of parameter combinations, the best-performing model on the validation set is further used to make predictions on the test data.  
In addition, the model iterates for 50 epochs in each experiment, and the model parameters corresponding to the epoch with the smallest loss on the validation set are used when conducting spatio-temporal forecasting on future data.

\subsection{Results and Analysis}
\label{sec:result}

We conducted numerical experiments on the Melbourne pedestrian counting dataset using four methods.
\begin{itemize}
   
    \item \textit{VAR}: the classic statistical vector autoregressive model.
    \item \textit{GRU}: the original GRU model, which does not incorporate the spatial correlation. 
  
    \item \textit{DCGRU}: DCGRU represents the classic DCGRU model that only relies on geographic information when building the sensor graph. 
    \item \textit{DCGRU-DTW}: The proposed DCGRU-DTW model incorporates both geographic information and time series information. 
\end{itemize}
All three neural networks, GRU, DCGRU, and DCGRU-DTW, are trained under the sequence-to-sequence framework introduced in Section~\ref{subsec:seq2seq}. 
The averaged experiment results are summarized in Tables~\ref{table1}--\ref{table3} with input lengths of 5 and 168 respectively. Different columns (1h, 2h, 3h, 4h, 5h) represents the number of hours in advance for making predictions.

\begin{table}[!ht]
    \centering
    \caption{Performance of different models for the various prediction intervals (1 to 5 hours) with $L_{input} = 5$}
    {\small
     \setlength{\tabcolsep}{3.75mm}{
    \begin{tabular}{ccccccc}
  \toprule
        \textbf{Model} & \textbf{Metric} & 1 h& 2 h& 3 h& 4 h& 5 h\\ \midrule
        ~ & MAE & 82.877  & 104.246  & 113.448
  & 118.342  & 121.832  \\ 
        VAR & MAPE & 31.58\% & 43.25\% & 52.68\% & 63.02\% & 70.63\% \\ 
        ~ & RMSE & 207.392  & 244.629  & 255.542  & 261.815  & 267.395  \\
    \hline
        ~ & MAE & 112.304  & 173.680  & 225.612  & 264.483  & 294.915  \\ 
        GRU & MAPE & 39.47\% & 71.49\% & 106.64\% & 160.94\% &  224.02\% \\ 
        ~ & RMSE & 237.811  & 334.622  & 429.453  &  507.356  & 554.252  \\ 
    \hline
        ~ & MAE & 80.534  & 109.435  & 131.126  & 147.224  & 157.427  \\ 
        DCGRU & MAPE & 27.39\% & 35.23\% & 41.42\% & 45.50\% & 46.91\% \\ 
        ~ & RMSE & 191.606  & 237.132  & 277.204  & 310.549  & 324.449  \\ 
    \hline
        ~ & MAE & 78.486  & 104.766  & 125.336  & 141.557  & 151.363  \\ 
        DCGRU-DTW & MAPE & 26.06\% & 32.77\% & 37.02\% & 41.15\% & 43.33\% \\ 
        ~ & RMSE & 190.672  & 234.447  & 271.101  & 306.298  & 319.305  \\ \bottomrule
    \end{tabular}}}
    \label{table1}
\end{table}

Table~\ref{table1} compares the performance of the four methods with the input length $L_{input} = 5$, {that is, using the past 5 hours' information for prediction}. First, we consider the three deep neural network methods of GRU, DCGRU and DCGRU-DTW. 
Compared with the classic DCGRU model, the average MAPE of DCGRU-DTW is around 1.3\% lower than that of DCGRU when predicting one hour in advance. Moreover, We notice that the advantage of DCGRU-DTW increases as the forecasting range extends. For example, when DCGRU-DTW makes 3 hours to 5 hours ahead prediction, the average MAPE of DCGRU-DTW is 3\% to 4\% lower than that of DCGRU. The RMSE and MAE achieved by DCGRU-DTW are also smaller than DCGRU when making 3 hours to 5 hours ahead forecasting: the MAE of DCGRU-DTW is around 6 units smaller, and the RMSE is around 4 units smaller, compared to DCGRU. 
In addition, it can be concluded that the classic DCGRU shows significantly better predictive capability than GRU. When making predictions 1 hour to 2 hours ahead, DCGRU reduces nearly half of the MAPE produced by GRU, and DCGRU's MAE is about 32 units smaller than GRU. DCGRU's RMSEs are also significantly lower. When predicting 3 hours to 5 hours in advance, the advantages of DCGRU and DCGRU-DTW over GRU increase. This could be associated with the fact that DCGRU's diffusion convolution operation can effectively capture the spatial dependency between different sensors, and utilizing this correlation would yield more accurate prediction results. The performance of VAR is not competitive in this experiment, indicating that traditional statistical models have difficulty in capturing complex spatiotemporal correlations of data compared to deep neural networks, and thus, they are unable to yield accurate predictions.

In order to take advantage of the daily and weekly periodicity of the data, we increase $L_{input}$ to 168, and we present the results by the three deep neural network methods in Table~\ref{table3}. For the VAR method, the best order is chosen by cross-validation, and the results are presented in Table~\ref{table1}. Compared to the result for $L_{input} = 5$,  the performance of GRU, DCGRU and DCGRU-DTW in each forecast range has witnessed significant improvement, especially when forecasting longer hours ahead. The performance of DCGRU-DTW is the best among the three methods. When forecasting 5 hours in advance and $L_{input}  = 168$, the MAE of DCGRU-DTW is approximately 5 units smaller than DCGRU and 30 units smaller than GRU. Furthermore, the MAPE of DCGRU-DTW is 3.5\% smaller than DCGRU and 25\% smaller than GRU, while the RMSE of DCGRU-DTW is 10 units smaller than DCGRU and 33 smaller than GRU. This shows the consistent goodness of DCGRU-DTW when changing the time length of the input data.

\begin{table}[htbp]
    \centering
    \caption{Performance of different models for the next $1$ to $5$ hours with $L_{input} = 168$}
    \setlength{\tabcolsep}{3.75mm}{
    \begin{tabular}{ccccccc}
    \toprule
        Model & Metric & 1 h& 2 h& 3 h& 4 h& 5 h\\ \midrule
        ~ & MAE &   79.288 &  103.376  &   118.899 & 129.827 &   136.224  \\ 
        GRU & MAPE & 27.91\% &   38.18\% &  47.97\% &  53.83\% &  56.22\% \\ 
        ~ & RMSE &   187.635 &   226.591  & 253.308   &  273.713 &  282.853 \\ \hline
        ~ & MAE & 72.812   &  89.781  &  100.697 & 107.623 & 111.761  \\ 
        DCGRU & MAPE &  26.66\% & 29.77\% & 32.33\% & 33.70\% & 34.80\% \\ 
        ~ & RMSE &  184.072  & 216.248 &  240.139  & 253.995  &  259.775  \\ \hline
        ~ & MAE & 72.232  & 87.635  & 97.162  & 103.256  & 106.853  \\ 
        DCGRU-DTW & MAPE & 25.81\% & 27.62\% & 28.72\% & 29.60\% & 31.30\% \\ 
        ~ & RMSE & 186.871  & 217.423  & 235.401  & 244.235  & 249.744  \\ \bottomrule
    \end{tabular}}
    \label{table3}
\end{table}

\begin{figure}[tbp]
    \centering
    \includegraphics[height =0.9\textheight]{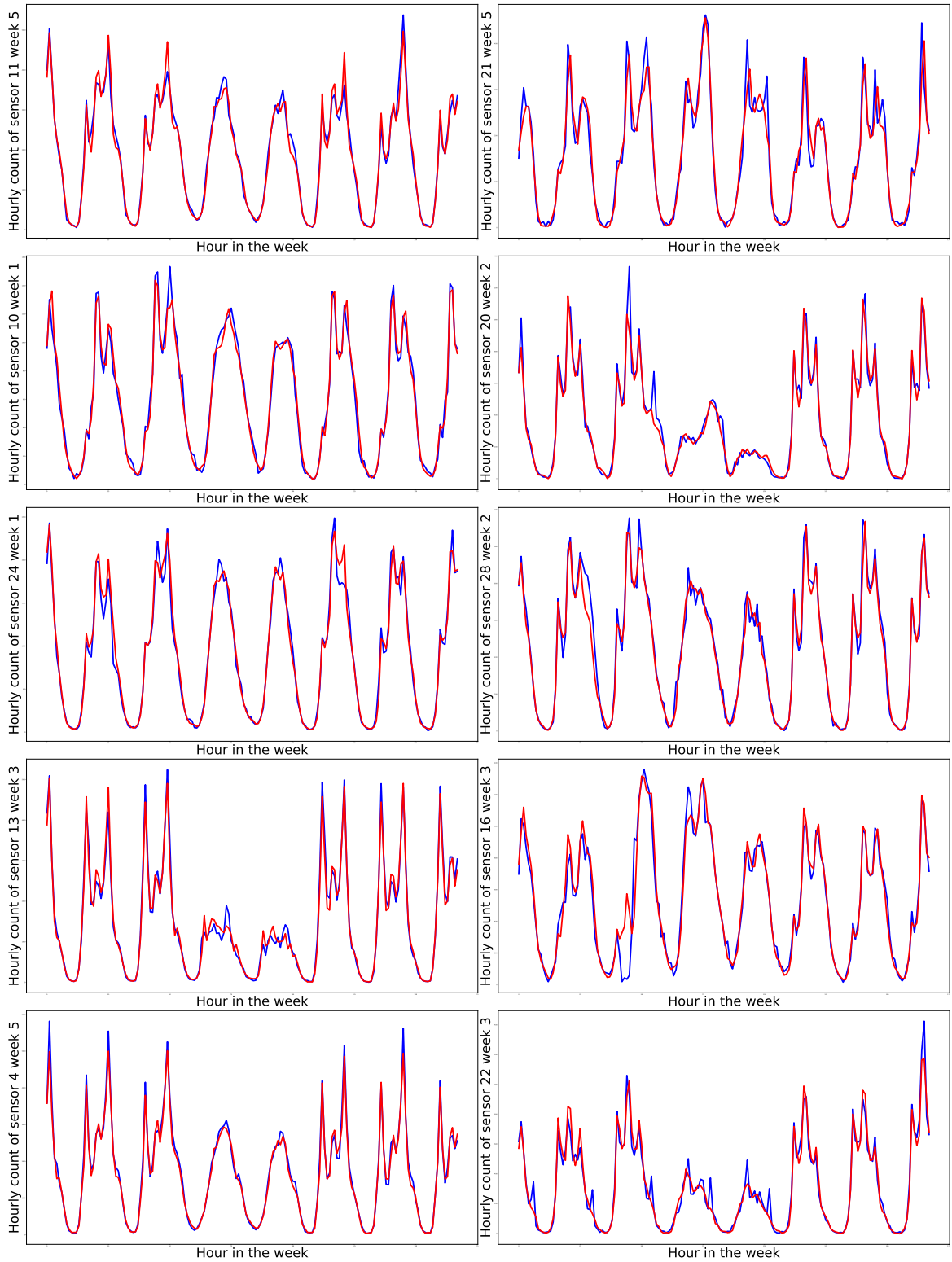}
    \caption{Ground truths (in blue) and $1$-hour-ahead prediction results (in red) of 10 selected sensors in different weeks}
    \label{fig:drawall}
\end{figure}

We also conduct a specific analysis of the prediction result of each sensor when DCGRU-DTW forecasts different hours ahead. For illustration, we randomly select 10 sensors and one-week data obtained from these sensors. Figure~\ref{fig:drawall} compares the true values of hourly pedestrian counts (in blue) and the predicted values by DCGRU-DTW (in red). It is observed that DCGRU-DTW can generally provide satisfactory predictions of pedestrian volume. DCGRU-DTW can capture the complex nonlinear trends in pedestrian volumes effectively, and in most cases, it can produce accurate predictions when the pedestrian volume presents drastic and oscillating changes within a short period of time, such as the data of sensor 4 in week 5 and the data of sensor 13 in week 3 (see Figure~\ref{fig:drawall}).

\section{Discussion and Conclusions} \label{sec:conclusion}

This study presents a diffusion convolution gated recurrent unit model with dynamic time warping for predicting pedestrian volumes at a city-wide scale. We used the data from the City of Melbourne pedestrian counting sensors to set up the experiments and evaluate the performance of the DCGRU-DTW model. Specifically, our model considers the static geographic relationship among sensors, as well as, the dynamic similarity between pedestrian counts over time. Experiments conducted in this study show that the proposed model outperforms the classic DCGRU model and the VAR model in terms of MAE, MAPE and RMSE. Furthermore, it is found that the prediction accuracy could be further improved by increasing the input length so that daily and weekly trends are captured, making it an effective tool to be used for transport and mobility based problems. In the future, We plan to explore more on real-time pedestrian management by taking advantage of the proposed DCGRU-DTW model.

\bibliographystyle{chicago}
\bibliography{main}

\newpage
\appendix

\section{The $k$-medoids Clustering}

To unveil the general spatio-temporal patterns of pedestrian volumes, we applied an anomaly detection algorithm based on $k$-medoids clustering to eliminate the influence of data in abnormal weeks. Compared with other clustering algorithms such as $k$-means, the $k$-medoids method is less sensitive to outliers due to the fact that $k$-medoids only allow existing data points to be cluster centres, which is beneficial for anomaly detection \citep{kaur2014k}. For example, when an outlier is assigned to a cluster, it would affect the mean value of its cluster in the $k$-means algorithm, resulting in a large deviation between the mean value and most of the data in the cluster. This may cause the true abnormal data to be mixed with normal data. In contrast, the $k$-medoids algorithm can reduce this adverse effect. The procedure of the $k$-medoids algorithm adopted for the multivariate time series data is summarized in Algorithm~\ref{alg1}.

\begin{algorithm}[htb]  
	\renewcommand{\algorithmicrequire}{\textbf{Input:}}
	\renewcommand{\algorithmicensure}{\textbf{Output:}}
	\caption{$k$-medoids for multivariate time series data}  
	\label{alg:Framwork}  
	\begin{algorithmic}[1]  
		\Require  
		The data $\mathcal{D}_j$ of week $j$, $j=1,2,\ldots, N$,  the number of clusters $k$, the distance function $a$, and the maximum iteration number $P$
		\Ensure Cluster centers and cluster results of $\mathcal{D}_j$, $j=1,2,\ldots, N$ 
        \State Compute the distance matrix $A = (a_{jj'})_{N \times N}$ between the data of each week
		\State Randomly select $k$ points as medoids
  \State Set iteration counter $p = 1$
		\State Calculate the distance from each data point to the medoids
  \State Assign each data point to its nearest medoids and then $k$ clusters are formed
 \For {each cluster in $k$ clusters}
 \State Choose one data point as the target and compute the sum of distances of all data points to the target until each data point in this cluster has been selected as the target. 
 \State Select the data point that minimizes the above sum of distances as the new medoids
 \EndFor
		\While {new medoids are not completely the same as the former medoids \textbf{and} $p \leqslant P$}
		\State Repeat steps $4$ - $9$
\EndWhile

\Return New medoids and nearest medoid of $\mathcal{D}_j$, $j=1,2,\ldots, N$ 
	\end{algorithmic}  
 \label{alg1}
\end{algorithm}  

Note that each data point $\mathcal{D}_j = [\boldsymbol{X}_{1,j}, \ldots, \boldsymbol{X}_{N,j}]$ in Algorithm~\ref{alg1} is a $168 \times 30$ matrix, where $\boldsymbol{X}_{i,j}$ denotes the time series of the $i$-th sensor for the $j$-th week. Before calculating the distance, we first re-scaled the dataset for each sensor, that is, for $t$ in the $j$-th week, 
\begin{equation}
X_{i,t} = \frac{X_{i,t} - \min{\boldsymbol{X}_{i,j}}}{ \max{\boldsymbol{X}_{i,j}} -  \min{\boldsymbol{X}_{i,j}}}.
\end{equation}
The distance between  $\mathcal{D}_j$ and $\mathcal{D}_{j'}$ is 
\begin{equation}
a_{jj'} = \sum_{i=1}^{N} \text{DTW}(\boldsymbol{X}_{i,j}, \boldsymbol{X}_{i,j'}), 
\end{equation}
which is the sum of DTW distances of each sensor between the $j$-th week and the $j'$-th week.

We determine the number of clusters, $k$, based on the Silhouette score \citep{ROUSSEEUW198753}. The Silhouette score is a measure of how well each data point fits within its assigned cluster and can be used to guide the number of clusters in a clustering algorithm. It measures both the compactness of a cluster and the separation between different clusters. A higher Silhouette score indicates that the data points are well-clustered and have a clear separation between the clusters.

The Silhouette score for a single data point $i$, which belongs to cluster $I$, is calculated as follows:
\begin{enumerate}
    \item Calculate the average distance between data point $i$ and all other data points in the same cluster. Denote this value as $a_i$: 
\begin{equation}
a_i=\frac{1}{\left|C_I\right|-1} \sum_{j \in C_I, j \neq i} a_{ij}.
\end{equation}
where $\left|C_{I} \right|$ is the number of data points in cluster $I$, the same cluster as ${i}$, and $a_{ij}$ is the distance between data points $i$ and $j$. If the single data point $i$ itself is a cluster, i.e.  $\left|C_{I} \right| = 1$, then the Silhouette score of $i$ is 0. 
\item  Calculate the average distance between $i$ and all other data points in the nearest neighbouring cluster. Denote this value as $b_{i}$:
\begin{equation}
b_i=\min _{K \neq I} \frac{1}{\left|C_K\right|} \sum_{j \in C_K} a_{ij}.
\end{equation}
\item  Calculate the Silhouette score for data point $i$ as:
\begin{equation}
s_i=\frac{b_i-a_i}{\max \left(a_i, b_i\right)}.
\end{equation}
\end{enumerate}
The Silhouette score for the clustering algorithm is then calculated as the average of the Silhouette scores for all $N$ data points:
\begin{equation}
\text{Silhouette score} =\frac{1}{N} \sum_{i=1}^N s_i.
\end{equation}
The Silhouette score ranges from $-1$ to $1$. A score close to $1$ indicates that the data points are well-clustered, while a score close to $-1$ indicates that they are poorly clustered and may have been assigned to the wrong clusters. A score close to 0 indicates that the data points may belong to multiple clusters. Therefore, supported with this information, $k$ with the largest Silhouette score is regarded as the optimal number of clusters. In practice, we run the $k$-medoids algorithm for $1,000$ times, using different random seeds for the initialization of medoids. The optimal number of clusters we found is $k=2$. 

We chose the clustering result with the highest frequency as the final clustering setting, based on which data from abnormal weeks are identified and deleted. 
We identified the outliers using Tukey's method \citep{Salgado2016}:
\begin{enumerate}
    \item Calculate distances between the data of each week and its nearest centre; 
    \item Obtain the lower quartile $Q_1$ and upper quartile $Q_3$ of the calculated distances for each center;
    \item Calculate the upper critical value $Q_3+q(Q_3-Q_1)$ and lower critical value $Q_1-q(Q_3-Q_1)$, where $q$ represents the hyperparameter of the method.
\end{enumerate}
In Tukey's method, $q$ is typically chosen to be $1.5$ or $3$, and a data point is regarded as an outlier if its distance to the nearest centre is either greater than the upper critical value or less than the lower critical value. Here, we chose $q=1.5$ and deleted the data point whose distance to the nearest centre is greater than the upper critical value. After the pre-processing stage, the dataset of Melbourne's pedestrian volume with  $5,712$ hours of records was obtained.
\begin{figure}[tbp]
    \centering
    \includegraphics[width = 0.567\textwidth]{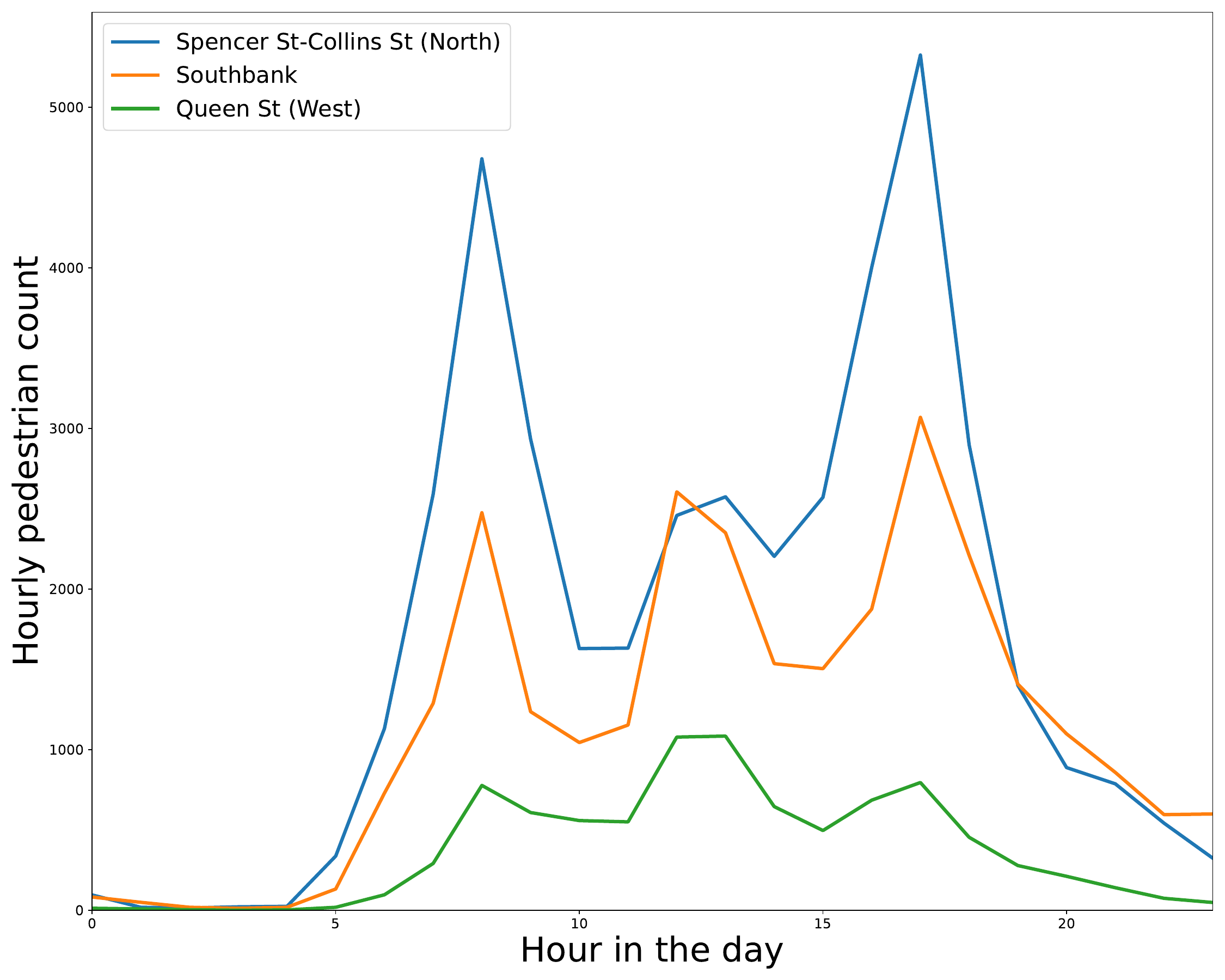}
    \caption{Pedestrian counts collected by three sensors in one day}
    \label{fig:south}
\end{figure}
\section{The Pedestrian Volumes at Three Locations}
 Figure~\ref{fig:south} shows the pedestrian volumes at three locations involving Sensor 13 (Spencer St-Collins St (North)), Sensor 19 (Southbank), and Sensor 20 (Queen St (West)) on one day. Despite Sensor 13 being geographically closer to Sensor 20, the time series of pedestrian volumes at Sensor 13 has a pattern more similar to that of Sensor 19.

\end{document}